# DIFFERENT APPLICATIONS OF MOBILE ROBOTS IN EDUCATION

Boris Crnokić[1], Miroslav Grubišić[2] and Tomislav Volarić[3]

[1,2]University of Mostar, Faculty of Mechanical Engineering and Computing, Mostar, Bosnia and Herzegovina.

[3] University of Mostar, Faculty of Science and Education, Mostar, Bosnia and Herzegovina.

***ABSTRACT***

*This paper presents different possibilities of using mobile robots in education. Through the application of mobile mechatronic robotic system "Robotino" this paper shows the possibilities of developing interactive lectures and exercises in order to raise the quality of education and to provide new competencies for students. Application of robot as a real system supports strengthening specific areas of knowledge and skills that the students develop through design, creation, assembly and operating with the robot. This way of learning contains a very important element and that is "Playful learning" or learning through play. Along with technical competences, combining this method with teamwork improves also social skills and motivation for learning. This paper will present application of the robot in education on examples of modelling and designing of mechatronic systems, simulating and parameters monitoring of mechatronic systems, and collecting, processing and application of data from sensors in mechatronic systems.*

## 1. INTRODUCTION

Mobile robots attract more and more attention of scientists because their range of use is increasing in various fields, such as: space exploration, underwater research, automotive, military, medicine, service robots, robots for serving in the home and entertainment robots, but also in all levels of education. Modern education systems have long been familiar with the term "educational robotics", i.e. learning environment in which people are motivated with creations that have characteristics similar to those of human or animal. Educational robotics is primarily focused on creating a robot that will help users to develop more practical, didactic, cognitive and motor skills. This approach is intended also to stimulate interest for research and science through set of different activities designed to support strengthening of specific areas of knowledge and skills that the students develop through design, creation, assembly and operating with a robot. Robotics is increasingly represented not only in higher education and scientific research institutions, but also in pre-school, primary and secondary education. [1]The company "Fischertecnhik" offers a multitude of projects and robotic systems that are used in education and training of children from 5 years up to the end of secondary school, from independent design and robotic systemstacking, to the programming of robot functions itself. [2] Perhaps the most common and best known of all educational robotic systems for all levels of education are those from the "LEGO" company, such as robot system "Lego Mindstorms NXT" and educational kit "LEGO WeDo". "Lego Mindstorms NXT" is a programmable robot that can be programmed through a large number of tools and has the possibility to use various sensors enabling students to learn about the different technical fields. "LEGO WeDo" is a simple set that allows students to construct different versions of robots and to program simple functions through simple software tools.[3] The Korean company "Robotis" produced an anthropomorphic robot "Darwin-OP" for education and research. This is a humanoid robot with advanced computer technology and sophisticated sensors, with large load capacity and ability of dynamic movement.[4] Previously mentioned robots are presented in Figure 1.

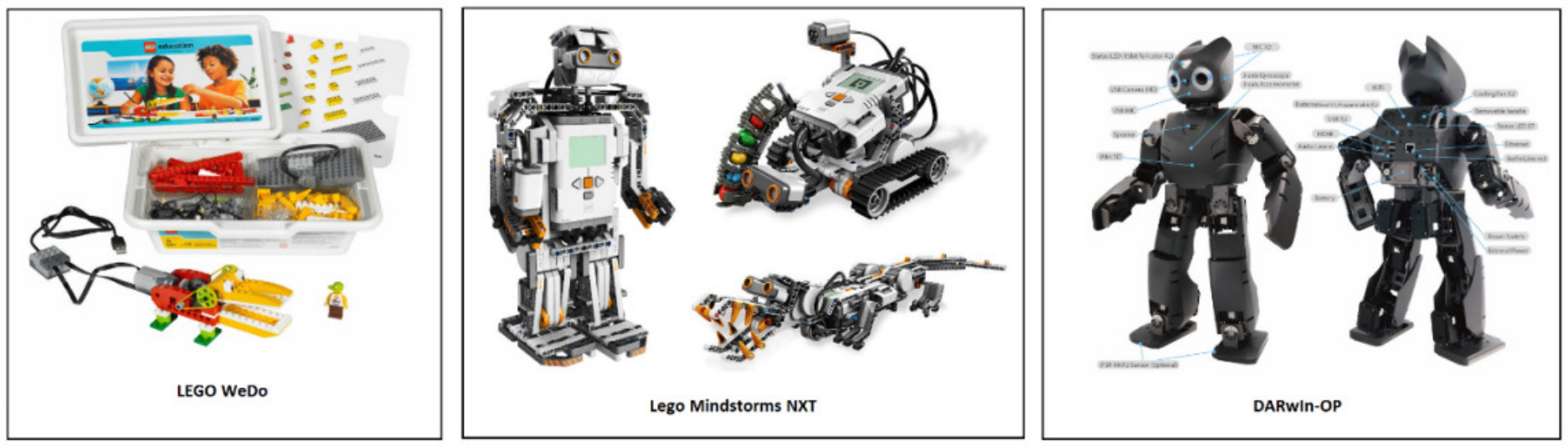

Figure 1.Some of the mobile robots that are applied in education

This paper will present application of mobile robot "Robotino" on various examples in education. The use of such a mobile robot system in education is very wide, in this paper we will present some examples of: learning about the components, structure and modelling of mechatronic systems, studying of automatic control, learning about the collecting, processing and application of data from different sensors, programming in different programming tools, studying methods of image processing and detection of objects.

## 1. HARDWARE AND SOFTWARE USED

Hardware used in this paper includes mobile robot Robotino 2 (Figure 2.) and mobile PC (laptop), while the application algorithms are implemented in programming environment *MATLAB/Simulink* and through software packages *RobotinoView2*. *Robotino SIM* simulating environment was used for the simulation of Robotino in the environment cluttered with obstacles.Mobile mechatronic system ***Robotino 2*** is manufactured by the German company *"Festo Didactic"*. Several series of this mobile robot system is produced and used for education, training, but also for the scientific research purposes. Robotino possesses various types of sensors, actuators and software interfaces which are at the highest level in the field of mobile robotics. Mobile robot system Robotino is applicable and adaptable for the secondary and university education in simple or more complex scientific projects, and through well designed training systems it is usable in lifelong learning programs. This fascinating and motivating system is used worldwide in educational and research institutions.

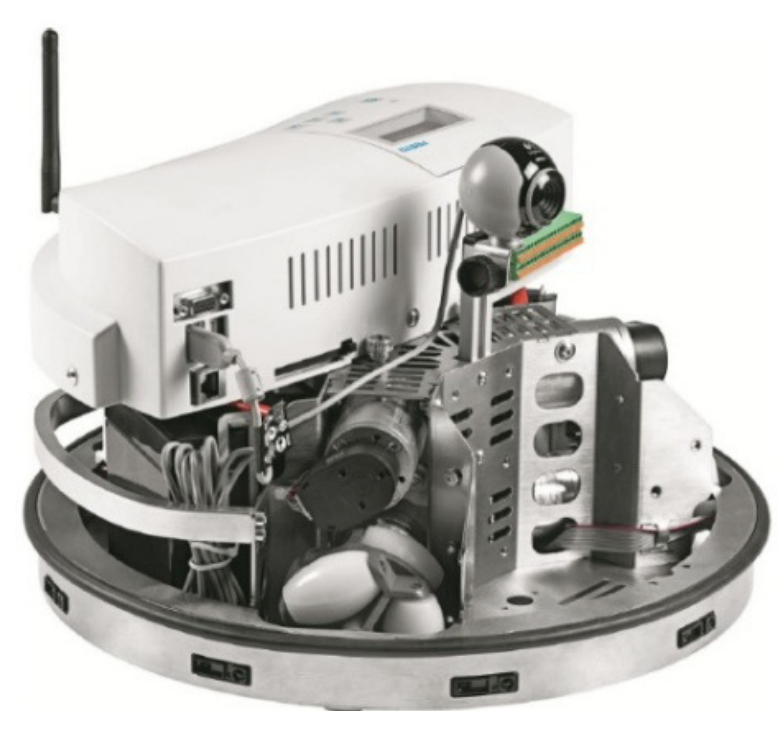

Figure 2. Robotino®2

There is a wide range of possibilities to create study programs or research projects with help of Robotino, and some of them are: collecting, scaling and analysis of data from a variety of sensors and its application for different functions and tasks; introducing with electric motor drive and control unit; study of closed control loop of mechatronic systems;graphical programming of applications for mobile robot; programming of mobile robot applications in programming languages such as .Net, C ++, C, C #, JAVA, or with Matlab or LabView programming interface; digital image processing and object tracking; indoor and outdoor robot navigation with obstacle avoidance; simulation of virtual applications in industrial production lines and laboratories (Figure 3.a); and many other applications in the field of mobile and service robotics. In addition to these application possibilities, Robotino is also the official equipment at the world competition in skills from different fields "WorldSkills International". Researchers from around the world are competing in the league of the "RoboCup Logistics League" in which Robotino is used to perform various tasks as an autonomous driverless operating system (Fig. 3.b). Company Festo organizes the competition in hockey "Festo Robotino Hockey Challenge", in which teams from different countries compete with teams composed of several Robotinos, and through playing the hockey they demonstrate skills and abilities to operate with mobile robots jointly in the groups.[5]–[7]

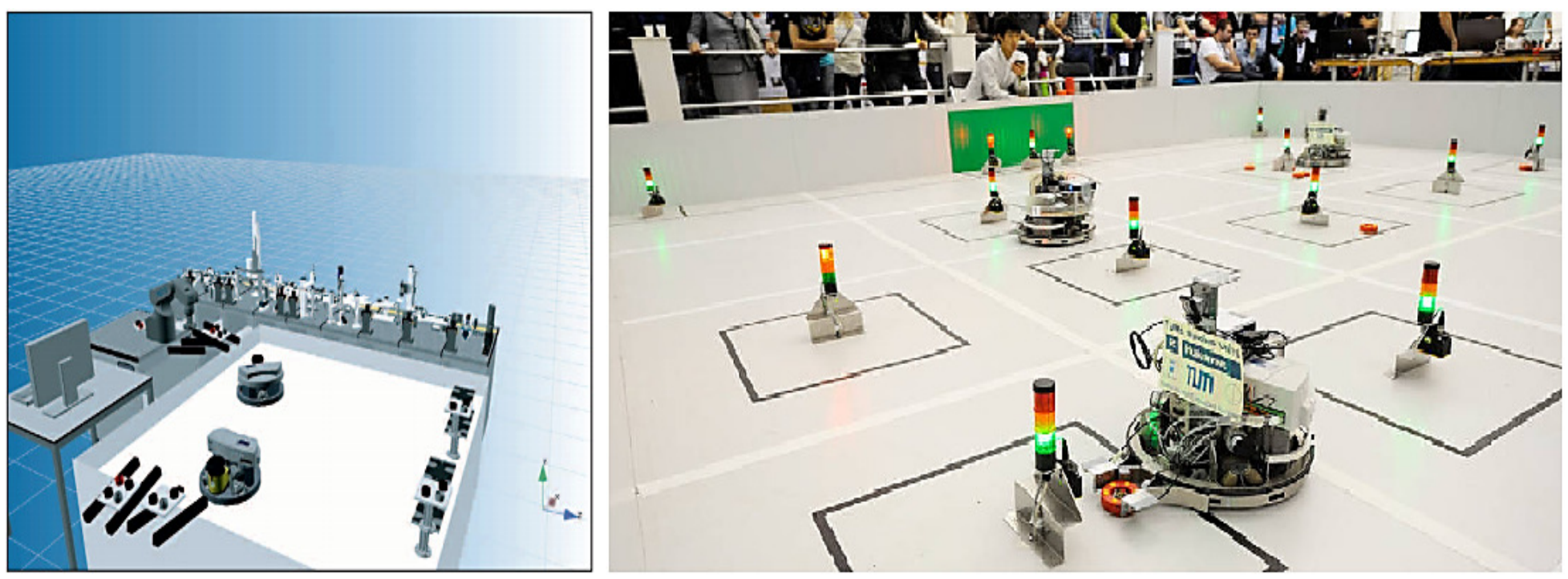

a) b)

Figure 3. a) Simulation of work in laboratories; b) Competition "RoboCup Logistics League"

***MATLAB*** (MATrixLABoratory) is a multifunctional numerical computational environment that allows manipulation of matrices, drawing of the functions, creating user interfaces and interfaces of programs written in other programming languages. Application of this programming environment is multiple, among other things, it is used to create programming algorithm for mobile robot motion control, as shown in this paper. ***Simulink*** is a graphical programming environment for modelling, simulation and analysis of dynamic systems. The primary interface of Simulink are tools of graphical block diagrams and flexible block-sets of different functions. Simulink is integrated with MATLAB, and allows integration of MATLAB algorithms in models of block diagrams and exporting of simulation results in MATLAB for further analysis.[8]

***RobotinoView2*** is an intuitive graphical programming environment designed specifically for the creation and implementation of control algorithms for mobile robot Robotino. This program combines modern operational concepts, upgrades by the user and intuitive use through a simple interface and a wide variety of functions,one of which is application of a visual sensor for detecting of different landmarks in the operating range of mobile robot.

***Robotino SIM*** is a Windows program for 3D simulation of Robotino in predefined virtual experimental environment. Robotino simulation model can not be further expanded, but it includes a geometrical model with three omnidirectional wheels, camera, nine distance sensors (IR sensors) and digital anti-collision sensor. This program allows control of Robotino with the use of RobotinoView2 or MATLAB programming interface. In the simulation environment is possible to create different scenes using various obstacles.

## 3. MATHEMATICAL MODELLING OF MECHATRONIC SYSTEMS

The main task of mathematical modelling of mechatronic systems is defining of models needed during the design of the optimized system structure that includes components such as sensors, actuators, comptrollers and mechanical components. Also, mathematical models are very important in the process of simulation and control of various functions within mechatronic systems. Mathematical modelling, essentially, represents theimplementation of mathematics to the real mechatronic system for examination, testing and validation of data that is later required to improve understanding and implementation of reliable mechatronic systems.In this case, students are introduced to mathematical modelling of mechatronic systems through the study of: *kinematic model, dynamic model, drive model and the robot control system model*.

By using the ***kinematic model*** and the speed data of the robot it is possible to determine position of the robot (Figure 4.). This modelling approach provides a simple mathematical procedure for assessing current position of the robot based on a predetermined position when speed and time of locomotion to the current positions are known.

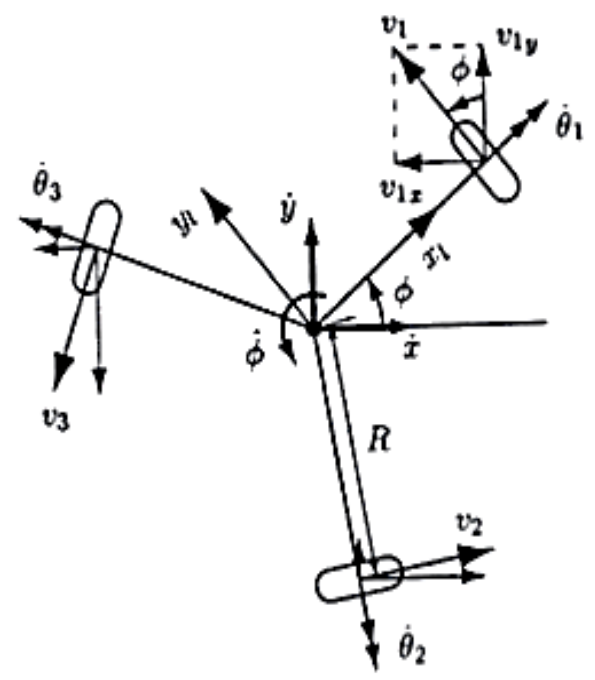

$$\omega^k = \begin{bmatrix}\omega_1(t)\\ \omega_2(t)\\ \omega_3(t)\end{bmatrix} = \frac{1}{r}\begin{bmatrix}v_1(t)\\ v_2(t)\\ v_3(t)\end{bmatrix} = \frac{1}{r}\begin{bmatrix}-\sin(\varphi_1) & \cos(\varphi_1) & R\\ -\sin(\varphi_2) & \cos(\varphi_2) & R\\ -\sin(\varphi_3) & \cos(\varphi_3) & R\end{bmatrix}\begin{bmatrix}\dot{x}_r(t)\\ \dot{y}_r(t)\\ \Omega(t)\end{bmatrix}$$

$$\begin{bmatrix}x_v(t_1)\\ y_v(t_1)\\ \theta(t_1)\end{bmatrix} = \begin{bmatrix}x_{v,0}\\ y_{v,0}\\ \theta_0\end{bmatrix} + r\int_{t0}^{t1}\left((T_{k\to v})^{-1}\begin{bmatrix}\omega_1(t)\\ \omega_2(t)\\ \omega_3(t)\end{bmatrix}dt\right)$$

Figure 4. Kinematic model of Robotino [9]

First step in design of the control structure of mobile mechatronic system is to determinate motion equations of the robot.In this case, ***dynamic model*** is essential to simulate locomotion of the robot, to analyse structure and construction of the robot and to optimize synthesis of control system. Figure 5. shows the simplified dynamic model of Robotino.

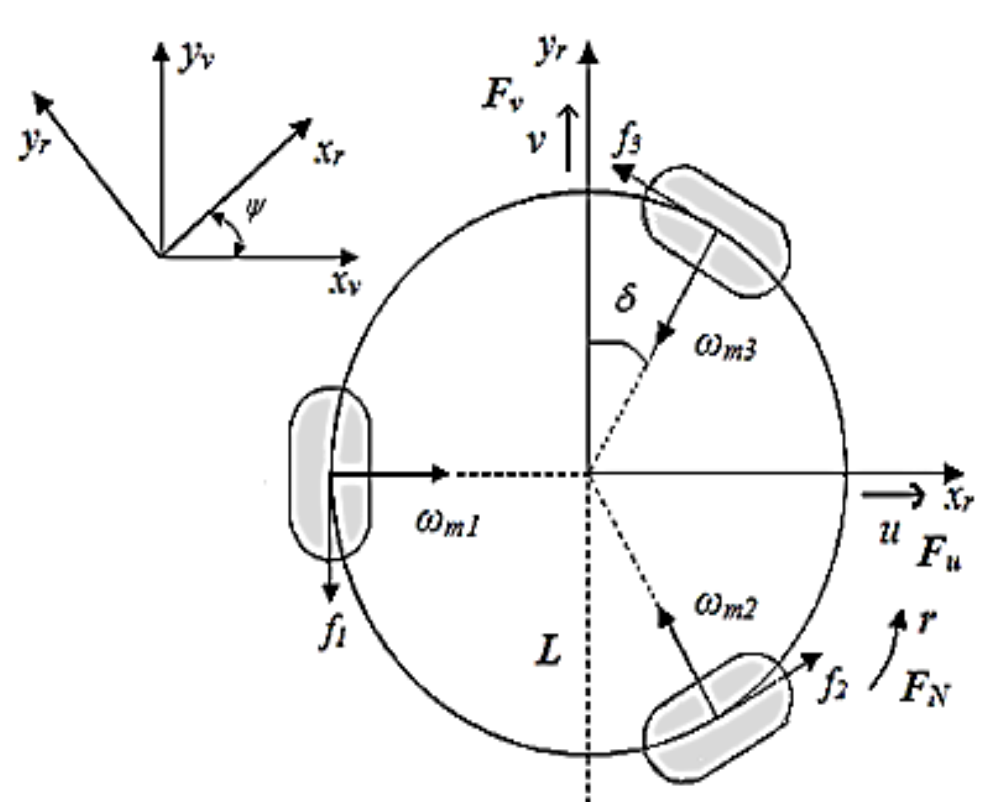

$$\begin{bmatrix}\dot{u}\\ \dot{v}\\ \dot{r}\end{bmatrix} = \begin{bmatrix}rv\\ -ru\\ 0\end{bmatrix} + \begin{bmatrix}\frac{1}{m} & 0 & 0\\ 0 & \frac{1}{m} & 0\\ 0 & 0 & \frac{1}{m}\end{bmatrix}\begin{bmatrix}F_u\\ F_v\\ F_N\end{bmatrix}$$

$$\begin{bmatrix}u\\ v\\ r\end{bmatrix} = (B^T)^{-1}\cdot\frac{R}{n}\begin{bmatrix}\omega_{m1}\\ \omega_{m2}\\ \omega_{m3}\end{bmatrix}$$

Figure 5. Dynamic model of Robotino[10]

The drive system of Robotino uses three DC motors. Using Kirchhoff's law it is possible to ***mathematically described models*** of each of the three motors (Figure 6.).

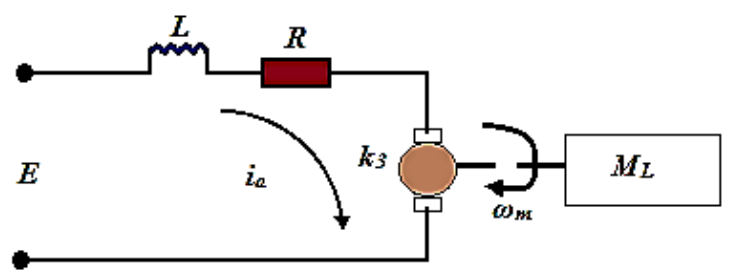

$$L_a \frac{di_a}{dt} + R_a i_a + k_3 \omega_m = E$$

$$J_0 \omega_m + b_0 \omega_m + \frac{Rf}{n} = k_2 i_a$$

$$\begin{bmatrix} \dot{u} \\ \dot{v} \\ \dot{r} \end{bmatrix} = G^{-1} \begin{bmatrix} rv \\ -rv \\ 0 \end{bmatrix} - G^{-1} HBB \left( \frac{k_2 k_3}{R_a} + b_0 \right) \frac{n^2}{R^2} \begin{bmatrix} u \\ v \\ r \end{bmatrix} + G^{-1} HB \frac{k_2 n}{R R_a} \begin{bmatrix} E_1 \\ E_2 \\ E_3 \end{bmatrix}$$

Figure6. Mathematical model of the motor

***Mathematical modelling of the control structure*** is very important because the control unit has the task to take the robot to desired position with desired orientation, along with gathering and processing input and output data. Mathematical model of the control system has two circuits with feedback loop. The main feedback loop is important for the position control of the robot, and other feedback loop uses infrared sensors to detect obstacles in the environment. Figure 7. shows a control block diagram of Robotino.

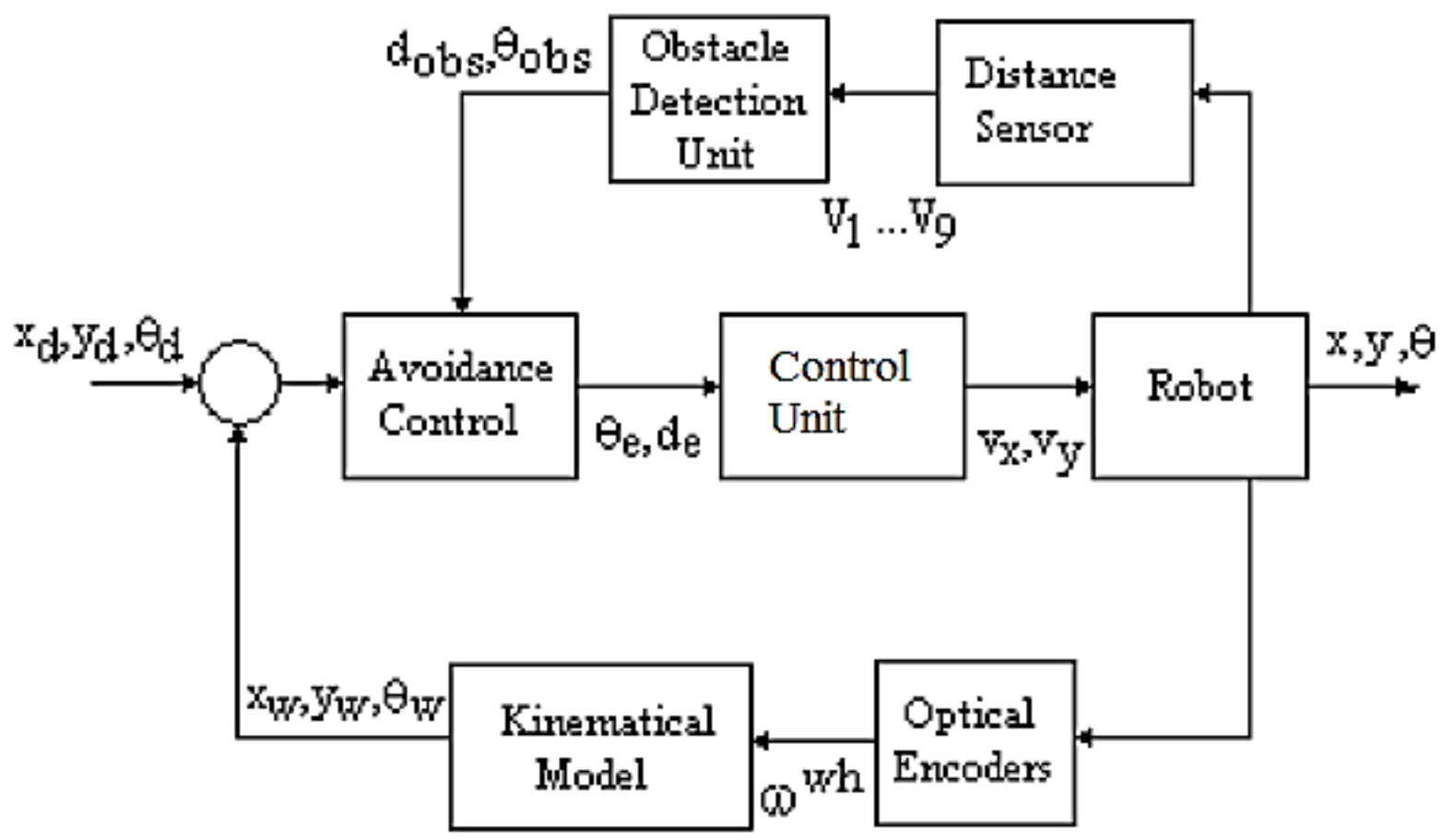

Figure 7.Control block diagram of Robotino

Mathematical modelling of mechatronic systems and application to the real system, in this case on the mobile robot, enables new ways of acquiring knowledge about mechatronic systems based on mathematical modelling and application of numerical methods. In this way it is possible to encourage students to study modelling of mechatronic systems in such a way as researchers do in a scientific work. That is, to enable students to acquire skills and knowledge in this field based on their own experience.

## 4. DESIGN OF MECHATRONIC SYSTEMS

Mechatronic systems are composed of four main groups of elements: *sensors, actuators, controllers and mechanical components*. Figure 8. shows a general model of mechatronic system within which all components of mechatronic mobile system Robotino are displayed. During the study of a general design of mechatronic systems, by using Robotino as an example, it is very simple and practical to study measuring and driving modules, communication and processing modules as well as modules of the output and feedback data. For easier understanding of designing any of mechatronic system, through work with Robotino, students can [7]:

- learn to manage an electrically controlled drive unit,
- learn the fundamentals, the construction, the measurement of values and theparameterisation of a direct current motor control

- learn the fundamentals of electrical drive technology,
- understand an omnidirectional 3-axis drive and know to operate it,
- are able to realise the commissioning of a mobile robot system using theRobotino as an example,
- move the Robotino in different directions,
- realise a sensor-guided path control,
- integrate image processing into the control of the Robotino,
- develop a sensor- guided autonomous path control of the Robotino usingobject recognition for example by colour,
- learn to mount and integrate of additional sensors into the Robotino system,
- mount and integrate handling devices into the system are able to program their own navigation and control algorithms using a C++librarycreate autonomous navigation of the Robotino.

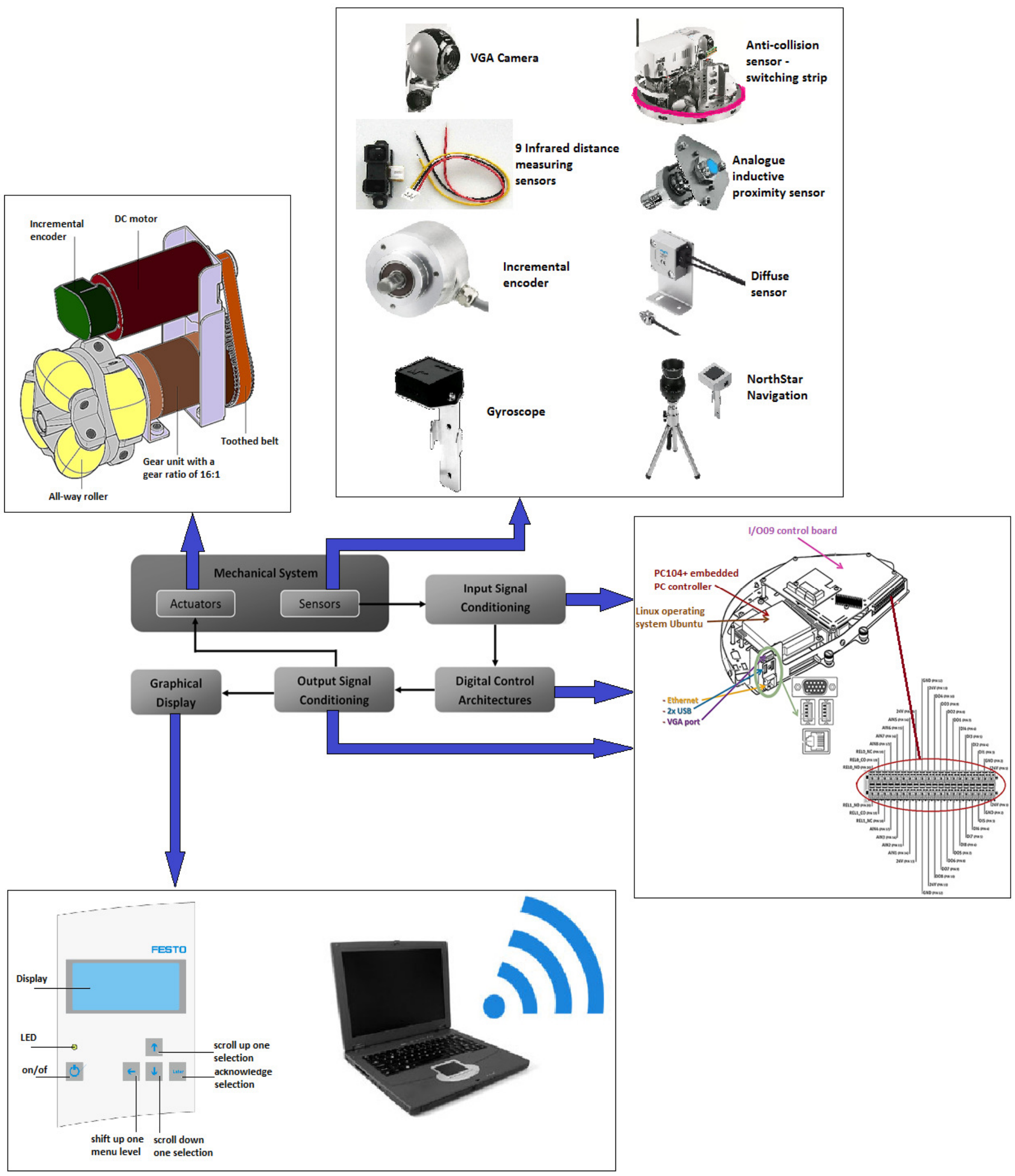

Figure 8. Components of mobile mechatronic system Robotino

## 5. Simulation Of Locomotion And Monitoring Of Mechatronic Systems Parameters

Modelling and simulation have a very important part in the realization of mechatronic system prototypes. Using the system such as Robotino students have chance to see integration of electrical components in mechanical model in order to examine and test functionality of mechatronic system. Respectively, through the analysis of simulation of locomotion and monitoring of system parameters, students can observe whether the system is physically optimally fused with I/O control structure, flexibility of system and measuring parameters.Prior to application of created algorithms it is possible to simulate all the movements and other tasks performed by Robotino by using programming interface of Robotino SIM (Figure 9.). By using diagnostic tool*EA09 View*it is possible to follow behaviour of all three motors, and to measure parameters such as speed, amperage, or the current position of the robot, and also to change the value of a PID controller, as shown in Figure 10. This diagnostic tool can be linked directly to the control unit of Robotino, or via Robotino View and MATLAB interface.

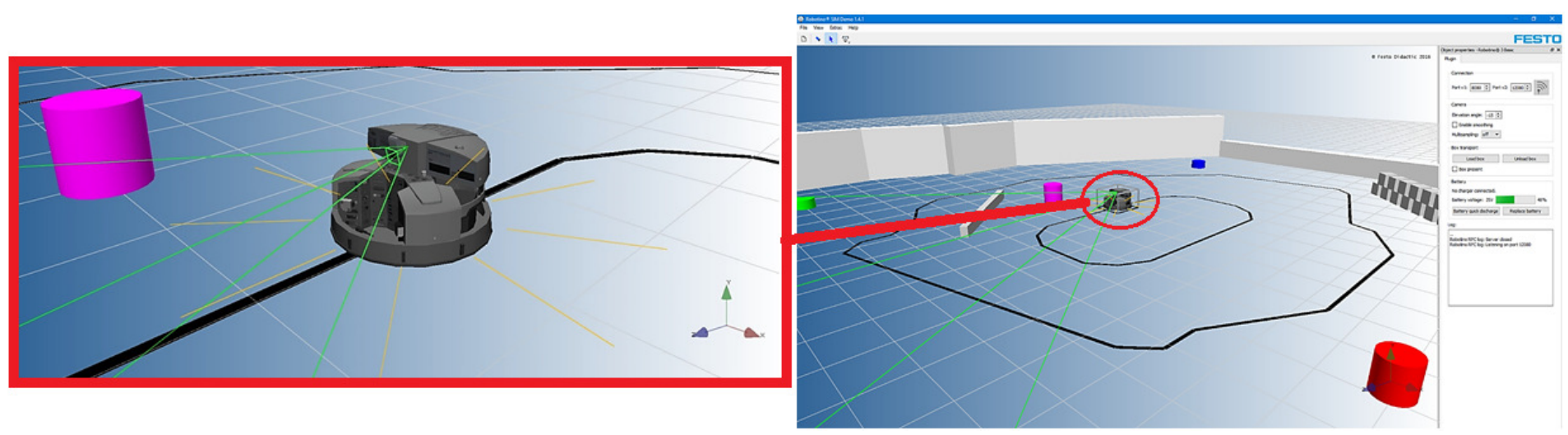

Figure 9. Robotino in simulation environment Robotino SIM

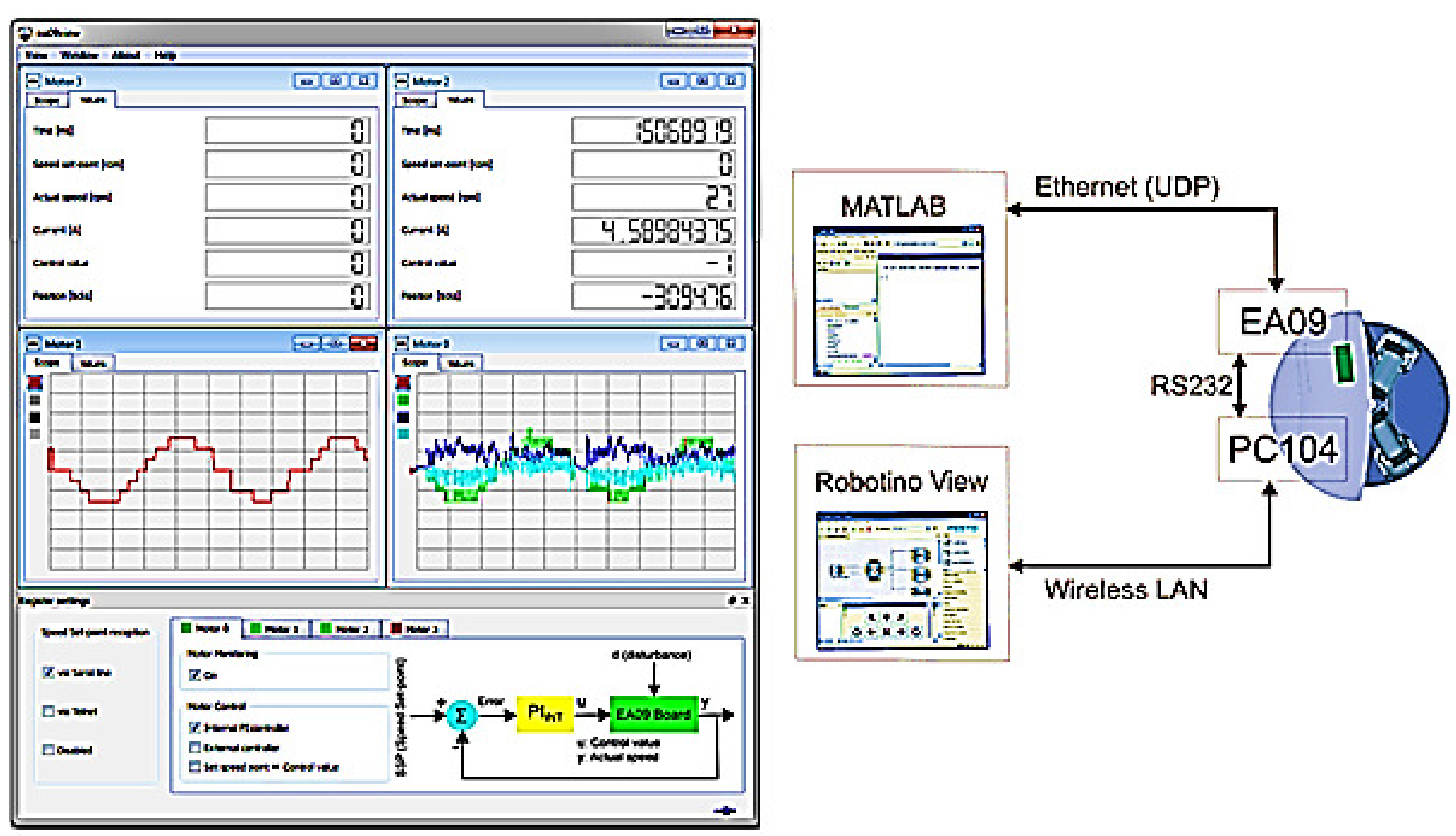

Figure 10. Use of EA09 View diagnostic tool for monitoring of the motor parameters

With application of these methods in the teaching process,by combining simulation and work on the real system, students develop the ability to understand different methods of modelling and simulation of mechatronic systems.Participating in exercises that are designed as a combination of simulation and real environment students are able to perform simple design and optimization tasks of mechatronic systems through the use of MATLAB/Simulink package, and to perform figuration of their models with different methods and toolboxes.

## 6. COLLECTING, PROCESSING AND APPLYING OF SENSORS DATA

One of the most important tasks of the mobile robot is the ability of collecting data about the environment in which it is located. This task can be performed by using different sensors whose measurements are being transformed ("translated") into useful informations for the control system of the robot. Sensors of mobile robots could roughly be divided into two groups: *sensors of the internal state* (which provide internal data about the movement of the robot): accelerometer, gyroscope, encoder, potentiometer, compass, etc.; and *sensors of the external state*(which provide external data about the environment): laser, infrared sensor, sonar, inductive sensor, camera, etc. Sensors of internal state provide information about the position of the robot in space, while the external state sensors are used for direct identification of places, obstacles and the situation in the environment.[11] By working with mobile robot Robotino students have the opportunity to see functionality of sensors of internal and external state through collecting, processing and applying of sensors data obtained in real application. This paper will show application of *vision sensor (camera)* and *infrared sensors (IR)* through exercises that are performed in the real environment and through simulation in the programming environment Robotino SIM.As an example of vision sensors application this paper showsan exercise of landmarks detection, where the lines on the floor are taken as landmarks. Depending on the detected landmarks mobile robot performs its movements and in this way it navigates through the space (room). This example was implemented in two different modes. For the first mode Simulink programming interface is used (Figure 11.), for second mode RobotinoView2 is used (Figure 13.).

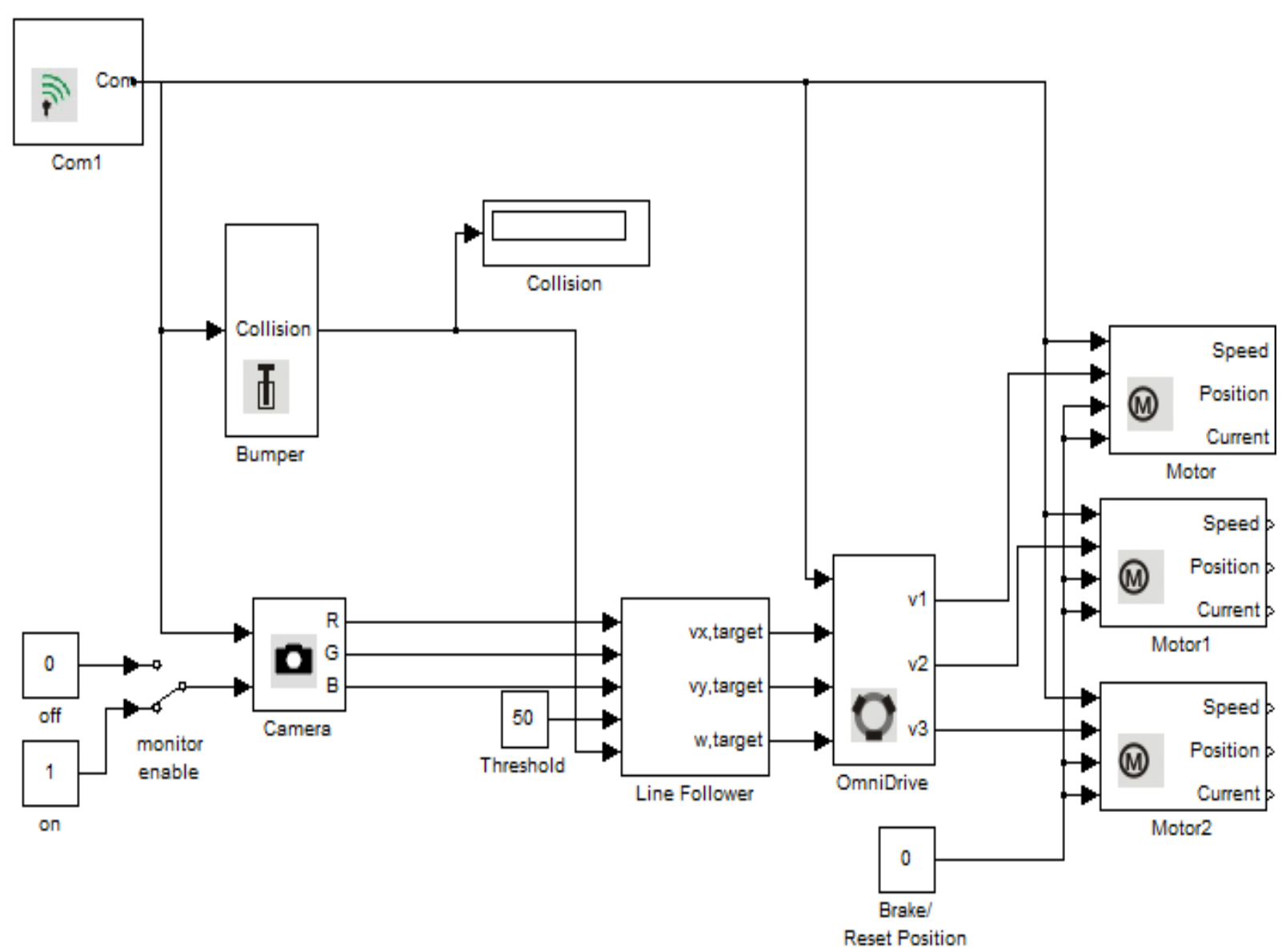

Figure 11.Simulink model of a system for line tracking

Pictures from the environment are collected through color camera on Robotino, while the processing is performed by using a Simulink block "Camera". After getting images in RGB values, images are sent to "Line Follower" block which detects edges of the line by using Prewitt edge detecting filter. When detection of the edges is completed, the block calculates $x$ and $y$ *position* of the line, $x$ and $y$ *velocity component* of the robot ($v_x$and $v_y$), and the angular velocity of the robot ($w$). Velocity data of the robot thereafter is sent to the "OmniDrive" block that controls all three motors and provides optimal line tracking. Figure 12. shows tracking of a line located on the floor using a Simulink model.

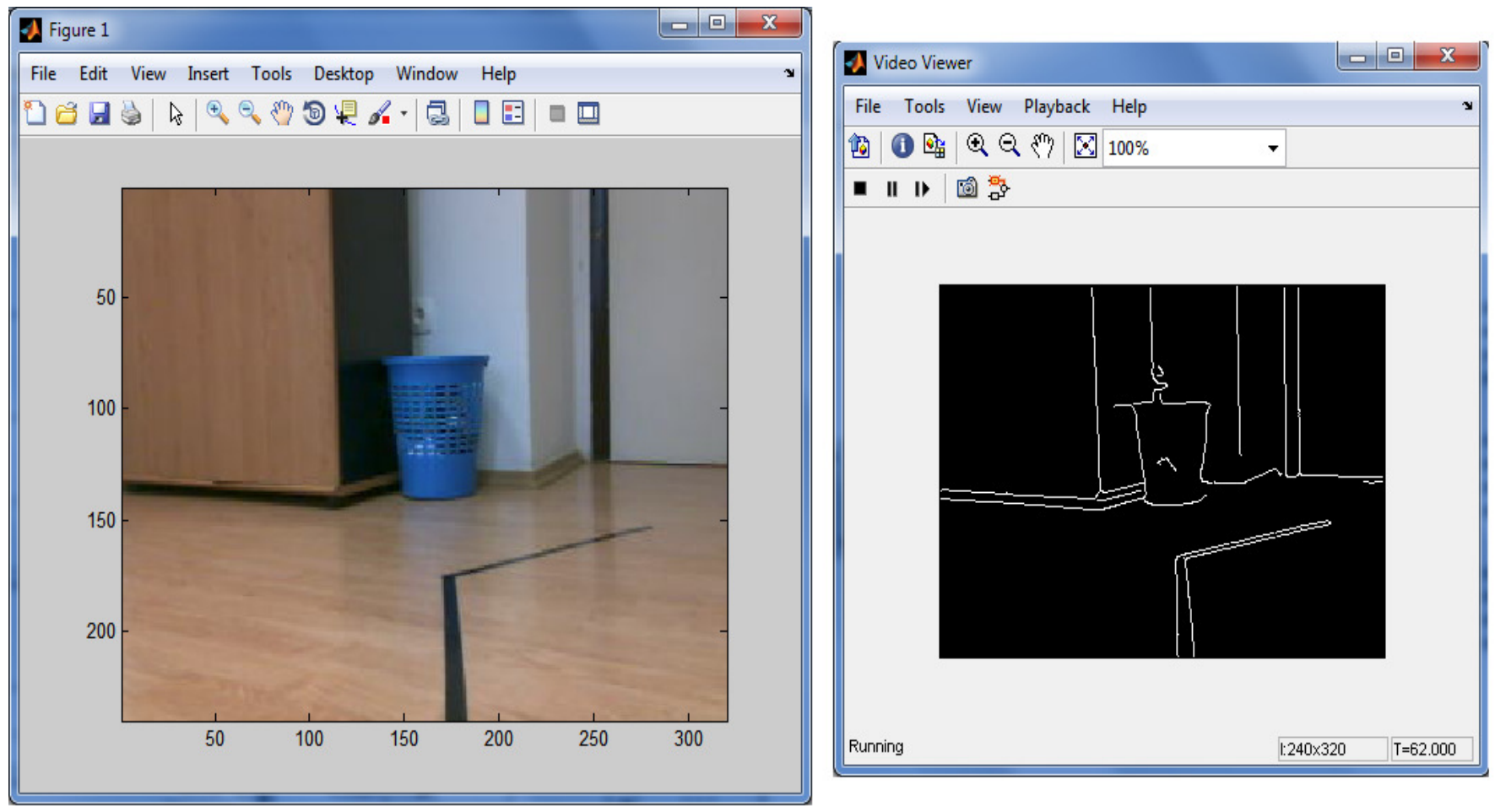

Figure 12. Line tracking on the floor (Simulink model)

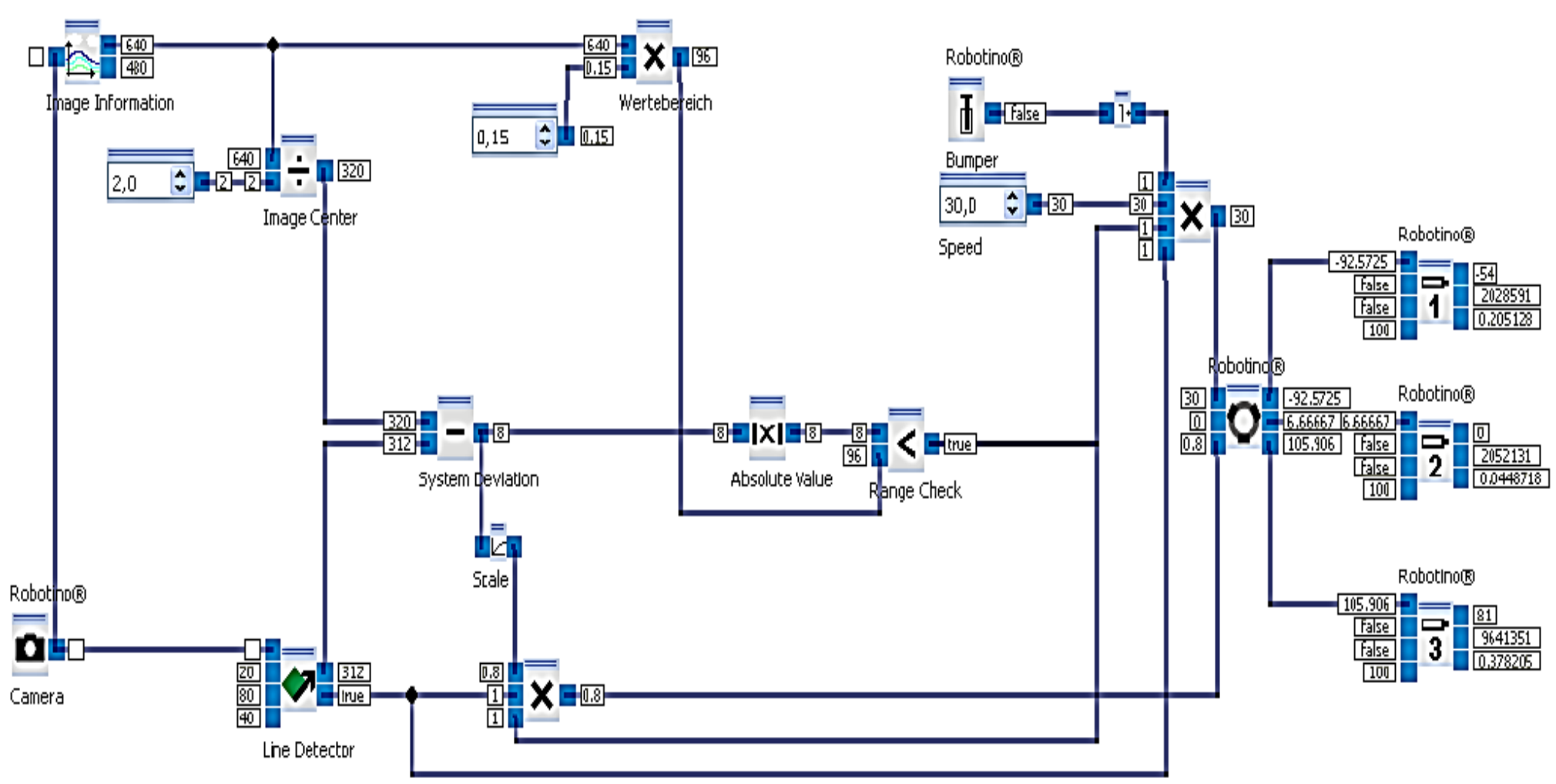

Figure 13. RobotinoViewmodel of a system for line tracking

In RobotinoView model Robotino sends compressed JPEG images that are decoded into RGB values. Detected lines are marked with a red cross on the image as shown in the example in Figure 14. This example also shows tracking of a line located on the floor.In order to enable tracking of lines it is necessary to define which output values are read out in the case of line detecting. X position of the detected line is read out from the X output. Left edge is specified as the value of "0", and the right edge is specified as the X value of image resolution (eg. 640 or 320), while the middle image is determined as half of the X value of image resolution (eg. 320 or 160). Robotino is rotating in a clockwise direction, if the sensor value is greater than 320 then Robotino rotates in the opposite clockwise direction. The range of values must be restricted to a narrow band around the middle of the field because high value of detected lines can result with excessive rotation of the robot in the left or right direction, so it may interrupt the tracking of the line or can stop the robot. Size of the area depends on the curviness of the path and the speed of the robot.[12]

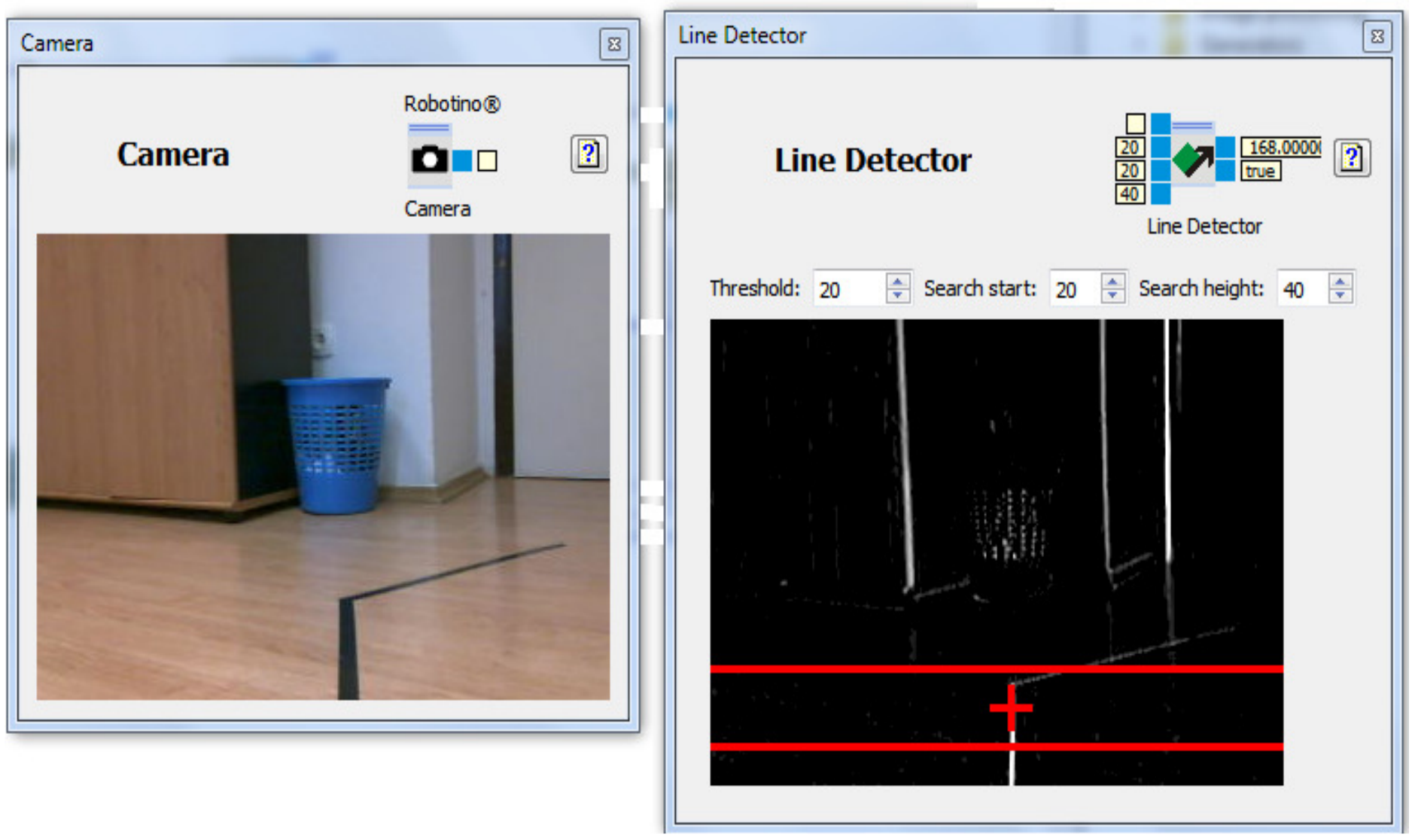

Figure14. Line tracking on the floor (RobotinoView model)

The following example shows application of IR sensors for detecting and avoiding of obstacles, based on the principle shown in Figure 15. This example is realized through a software package MATLAB, while detection and avoidance of obstacles is simulated in Robotino SIM simulation environment (Figure 16). Through such examples, students have the opportunity to study principles of distance sensors, data collection methods and A/D conversion, and ultimately to see the application of the data obtained from the sensors to detect obstacles in the environment of mobile robot.

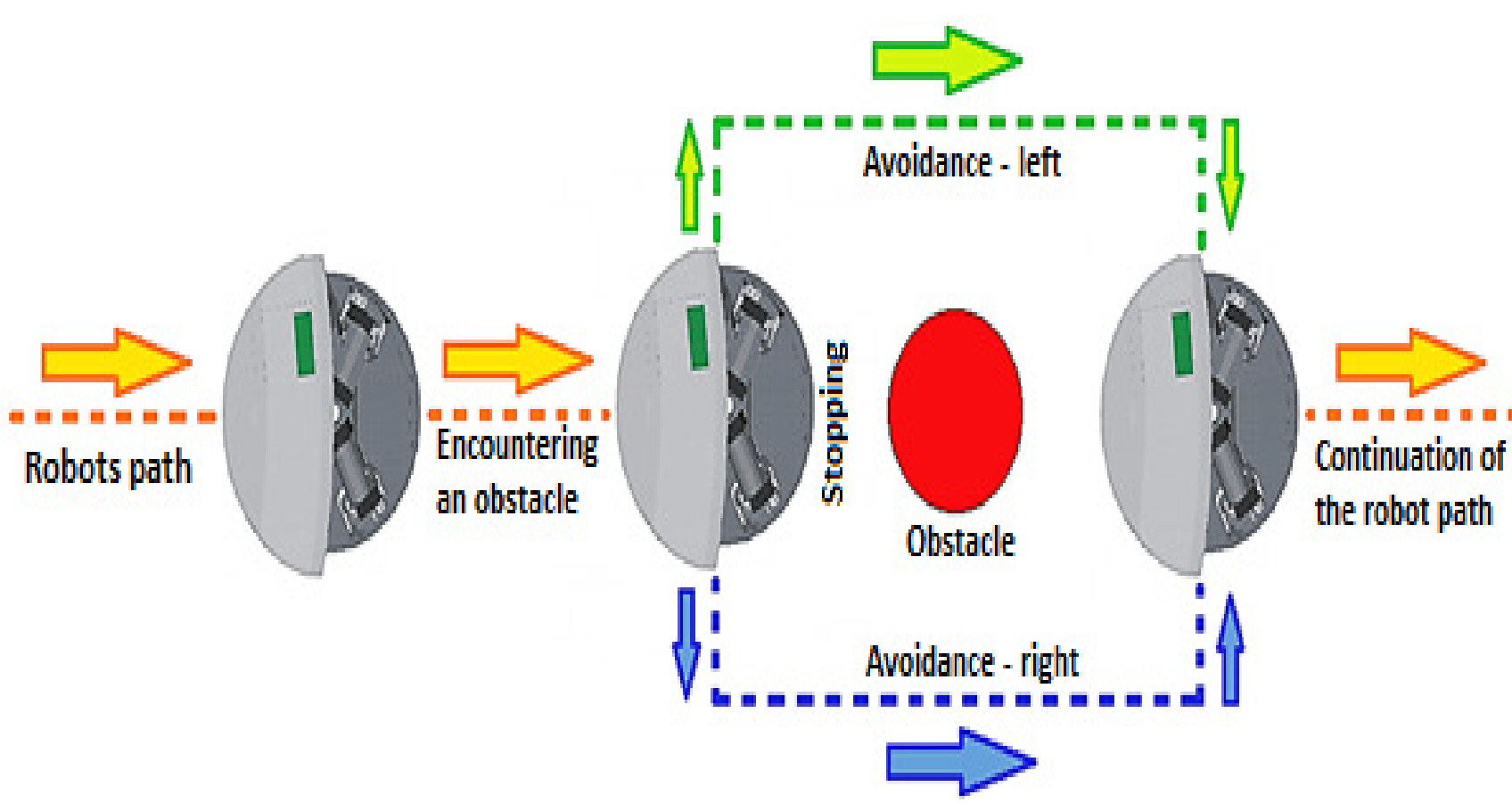

Figure 15. Avoidance of detected obstacles (working principle)

MATLAB example illustrates the use of the Distance Sensors on Robotino. If Robotino is placed in an environment with obstacles, then it moves around avoiding the obstacles. Robotino is equipped with 9 infrared distance sensors (labelled IR0-IR8 on Figure 17.a). In this example, we take the voltage readings from the 0th (IR1), 1st (IR2) and 8th (IR9) sensor as they are in front of the robot (Figure 17. b). Based on the readings from three distance sensors, we can find out if Robotino is approaching an obstacle or not. In case it is, Robotino stops and rotates until all the three distance sensors do not detect any obstacles in front of them. If the sensors don't detect any obstacles in front of them then Robotino moves forward with a constant velocity. The example runs for 60 seconds.

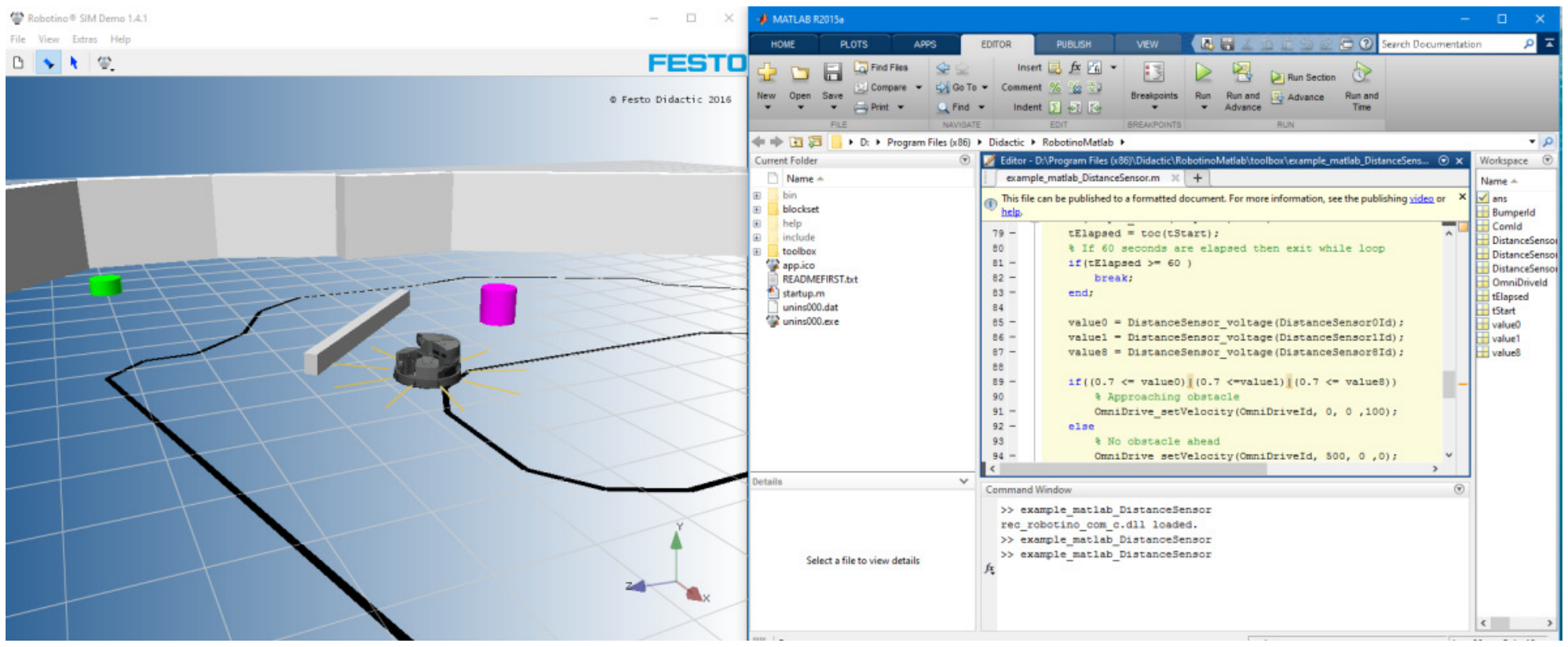

Figure 16. Application of MATLAB and Robotino SIM

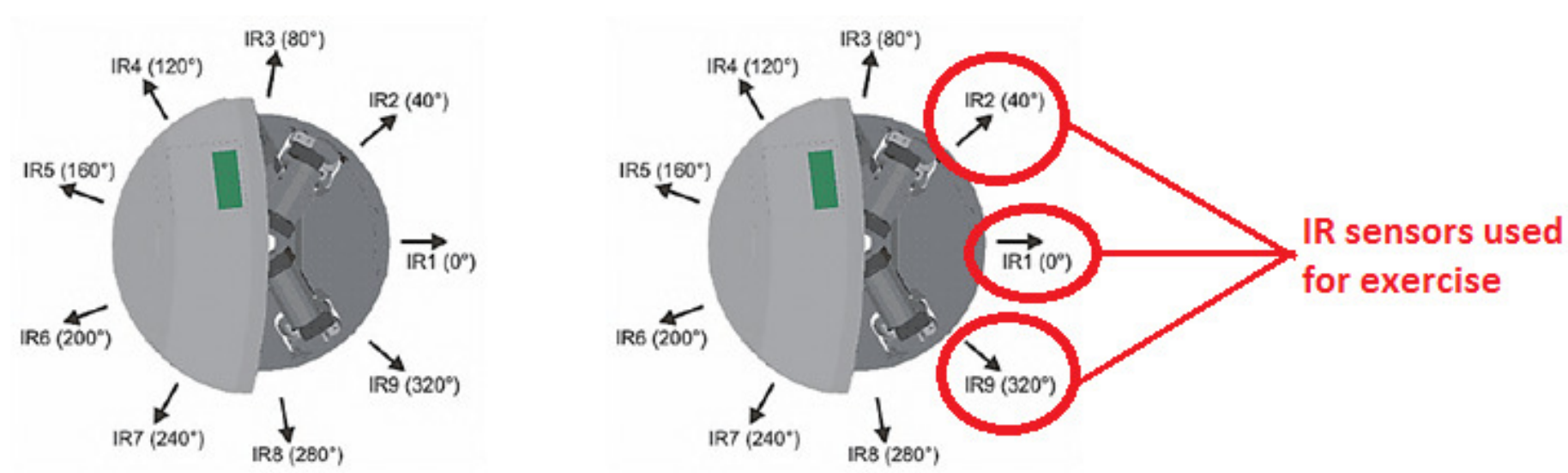

a) b)

Figure 17. IR sensors on Robotino

MATLAB code for this exercise is shown below:

```
ComId = Com_construct;
OmniDriveId = OmniDrive_construct;
DistanceSensor0Id = DistanceSensor_construct(0);
DistanceSensor1Id = DistanceSensor_construct(1);
DistanceSensor8Id = DistanceSensor_construct(8);
BumperId = Bumper_construct;

Com_setAddress(ComId, '127.0.0.1:8081');
Com_connect(ComId);

OmniDrive_setComId(OmniDriveId, ComId);
DistanceSensor_setComId(DistanceSensor0Id, ComId);
DistanceSensor_setComId(DistanceSensor1Id, ComId);
DistanceSensor_setComId(DistanceSensor8Id, ComId);
Bumper_setComId(BumperId, ComId);

tStart = tic;
while (Bumper_value(BumperId) ~= 1)
tElapsed = toc(tStart);
% If 60 seconds are elapsed then exit while loop
if(tElapsed>= 60 )
break;
end;

    value0 = DistanceSensor_voltage(DistanceSensor0Id);
```

```
    value1 = DistanceSensor_voltage(DistanceSensor1Id);
    value8 = DistanceSensor_voltage(DistanceSensor8Id);

if((0.7 <= value0)|(0.7 <=value1)|(0.7 <= value8))
% Approaching obstacle
OmniDrive_setVelocity(OmniDriveId, 0, 0 ,100);
else
% No obstacle ahead
OmniDrive_setVelocity(OmniDriveId, 500, 0 ,0);
end;

end;

Com_disconnect(ComId);
DistanceSensor_destroy(DistanceSensor0Id);
DistanceSensor_destroy(DistanceSensor1Id);
DistanceSensor_destroy(DistanceSensor8Id);
Bumper_destroy(BumperId);
OmniDrive_destroy(OmniDriveId);
Com_destroy(ComId);
```

## 7. PEDAGOGICAL BENEFITS

Educational robotics is based on the use of robotics for creation of competencies as part of regular curricula, and involving students to study subjects with a new interactive approach.For the successful use of robotics as a learning tool for education, it is important to know that an essential role plays an interaction and a kind of “student’s playing”. Success of use robotics in educational process depends on elements that promote student’s engagement and at the same time pushes toward indifference with it.Use of robotics for easier mastering of certain curriculaencompassesstudent motivation for guided discovery and problem solving, and encourages students to teamwork and common finding of solutions. Such a way of curricula organisation promotes inclusion, and enables collaboration of talented students and students with learning difficulties. Educational robots fall into new educational technologies to facilitate learning and improve education, they are new technology tool that has the ability to improve and facilitate the education process.

Educational robots can be applied in different levels of education, from pre-school, elementary school, secondary school touniversity level. Education robots have to adapt to age, interest, previous knowledge and must stimulate the interest of children or student. Each age level requires special robots, with custom look, function, mode and software.The role of educational robotics should be seen as a means of stimulating important life skills (cognitive and personal, teamwork) through which students can develop their potential by using imagination, anddevelop research skills, creative thinking, decision-making, problem solving, communication and teamwork, adopting and developing skills in new technologies.

The use of robotics in education is still not very widespread, so it is difficult to get feedback from students and professors.This is a process that needs to be constantly developed and devised, and strategies for involving a large number of students into robotic education are already being offered.It is still unknown whether all the reasons for the introduction of robotic education have been achieved. When applying robots in education at our faculty, progress is made on the level of student interest in developing joint projects through team work. Students have more motivation for active participation in the teaching process, and show a certain level of easier understanding of teaching unitsin subjects related to sensors, actuators and robot programming.Students are actively involved into learning/teaching processes trough observation, experimentation, abstraction and theorization. They have apparently become aware of what the process of learning/teaching involves, and how to make the collection/transfer of knowledge more creative and simpler.

## 8. CONCLUSIONS AND RECOMMENDATIONS

The use of such mobile mechatronic systems meets a number of different training and vocational requirements. Robotino facilitates industry-orientated vocational training with the hardware that consists of didactically suitable industrial components.Through a well-designed lessons and exercises with this system, students acquire the following competencies:

- Social competence,
- Technical competence and
- Methodological competence.

As can be seen from the examples in this paper, the application of this system in education is wide and includes the following areas:

- *Mechanics* (mechanical construction of a mobile robot system),
- *Electrics* (control of drive unit, wiring of electrical components),
- *Sensors* (sensor-guided path control, path control by image processing of webcam pictures, collision-free path control with distance sensors),
- *Feedback control systems* (control of omnidirectional drives),
- *Use of communication interfaces* (wireless LAN),
- *Commissioning* (Commissioning of a mobile robot system).

Finally, the application of a complex system such as a mobile robot Robotino greatly simplifies for the students:

- understanding of the basic principles of mechatronic design,
- study of the components and interfaces of mechatronic systems,
- creation of application software,
- simulation and real time operation,
- application of mechatronics in the manufacturing process,
- application of mechatronics in transport systems and electric vehicles.

Study of mechatronic systems through this form of teaching, using the mobile robot Robotino, is very simple, and most importantly it is possible to practically show all dependencies and system performance through play, entertaining exercises and motivating teamwork. This kind of educational models allow students to set their own strategies to solve problems, and thus to indirectly acquire experience about the behaviour of the observed system and the possibilities for its modification.

## AUTHORS

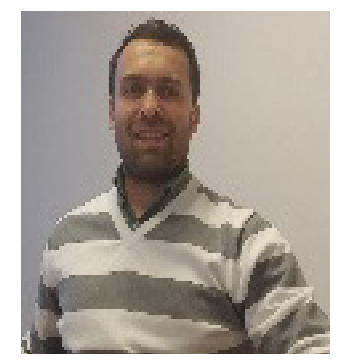

**Boris Crnokić** received the M.Sc. degree in mechanical engineering from University of Mostar,Faculty of Mechanical Engineering and Computing,Bosnia and Herzegovina, in 2008. He received Ph.D.degree in mechatronics and roboticsfrom University of Mostar, Faculty of Mechanical Engineering and Computingin 2016. He is presently Assistant ProfessoronFaculty of Mechanical Engineering and Computing. His research interests are mechatronics, robotics,multisensorsystems, robot navigation and engineering education. Dr. Crnokić is author and co-author of several papers published in international journals and conferences. He has participated in several national and Europeanresearch projects.

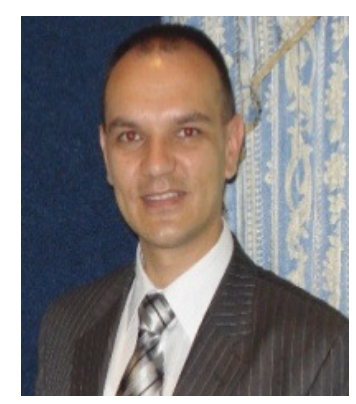

**Miroslav Grubišić** received the M.Sc. and the Ph.D. degrees in mechanical engineering from University of Mostar, Faculty of Mechanical Engineering and Computing, Bosnia and Herzegovina, in 2010. and 2017., respectively. Currently he is an Associate Professor of the Faculty of Mechanical Engineering and Computing on University of Mostar, and he is also director of the Institute of Mechanical Engineering in Mostar.Dr.Grubišić has been involved in several national and European projects. His research interests include automotive mechatronic systems, engines and motor vehicles, multisensor systems and engineering education.

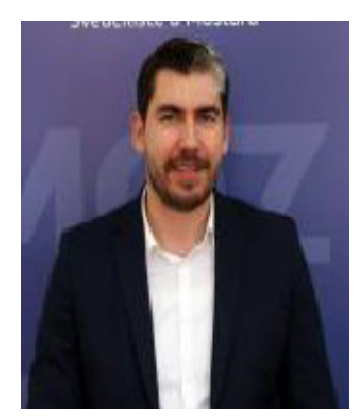

**TomislavVolarić** received the M.Sc. degree in computer engineering in 2010. from the University of Dubrovnik (Croatia) and a Ph.D. degree in computer sciencesfrom the University of Split (Croatia) in 2017. He has been involved in several national and European research projects and he is a Cisco certified network associate. He has ACM and IEEE membership since 2012. Dr.Volarić currently is an Associate Professor of computer engineering at the University of Mostar, Faculty of Science and Education,and his current research is focused on the development of eLearning systems.